# Assessing the Search and Rescue Domain as an Applied and Realistic Benchmark for Robotic Systems

Frank E. Schneider and Dennis Wildermuth
Cognitive Mobile Systems (CMS)
Fraunhofer Institute for Communication, Information Processing and Ergonomics FKIE
Wachtberg, Germany
{frank.schneider, dennis.wildermuth}@fkie.fraunhofer.de

*Abstract*—Aim of this paper is to provide a review of the state of the art in Search and Rescue (SAR) robotics. Suitable robotic applications in the SAR domain are described, and SAR-specific demands and requirements on the various components of a robotic system are pictured. Current research and development in SAR robotics is outlined, and an overview of robotic systems and sub-systems currently in use in SAR and disaster response scenarios is given. Finally we show a number of possible research directions for SAR robots, which might change the overall design and operation of SAR robotics in the longer-term future. All this is meant to support our main idea of taking SAR applications as an applied benchmark for the Field Robotics (FR) domain.

## I. Introduction

All over the world natural disasters and crisis situations, such as earthquakes, tsunamis, hurricanes or resulting nuclear catastrophes, pose an unpredictable yet significant risk to the lives and prosperity of the population. The ability of first response search and rescue teams to safely and efficiently provide aid in the inherently harsh environments of disasters means a significant challenge. Several disasters in the last ten years (Fukushima nuclear power plant, Katrina hurricane, Tohoku tsunami and earthquake) have underlined the need for robotic platforms able to assist Search and Rescue (SAR) operations in scenarios which are hazardous for human personnel to enter.

Since the early 2000's robotic solutions have been utilised in many response efforts in disaster incidents, demonstrating their potential to reduce the risk of loss of life, accelerate response times and gather essential data [1]. Starting in 2005, the first use of small unmanned aerial vehicles (UAV) could be seen. Later, aerial vehicles became the standard tool in disaster scenarios. In fact, since 2011, only one disaster did not use a UAV and that was the South Korea ferry where they used an underwater vehicle. So the field moved from being ground robots dominated (pre 2005) to shifting toward unmanned aerial vehicles. In about 2007, it became also much more commonplace to see underwater vehicles being used. By now several projects have been funded such as the ICARUS project [2] with the overall purpose of applying innovations for improving the management of a crisis and by doing so to reduce the risk and impact of the crisis on citizens.

The Center for Robot-Assisted Search and Rescue (CRASAR) at Texas A&M University has participated in 15 of the 35 documented deployments of disaster robots throughout the world and has formally analysed nine others, providing a comprehensive archive of rescue robots in practice. Their robots and protocols have been successfully used during the Hurricane Katrina emergency and the Fukushima nuclear accident and have been adopted by Italian and German governments [3].

Robots can be employed to search for casualties, provide them with first aid and essential goods, can map disrupted areas and assess structural damages to buildings. This shows the wide range of possible robotics applications during typical SAR deployments. Robot systems and all their involved components have to work absolutely at the cutting edge of robotics developments to tackle the pictured challenges. Thus, it appears fully reasonable to take SAR scenarios as realistic and applied benchmarks for research and development in the Field Robotics (FR) domain.

However, despite the high level already reached in robotics research in general, there is still significant room for improvement in search and rescue robotics, especially in the capabilities of robots to aid in the 'rescue' component of responses. Real world disaster response scenarios present challenges which need to be carefully investigated and rigorously solved in order to reach systems which really can assist humans during natural and man-made disasters.

In the following section II we describe typical areas of SAR applications: environment monitoring, disaster inspection, and the actual Search and Rescue task. In section III we picture SAR-specific demands and requirements on the various components of a robotic system. Section IV presents a state-of-the-art overview of robotic systems and sub-systems currently in use in SAR and disaster response scenarios. Finally, in section V we give an outlook on expected future robotic developments in the SAR domain.

## II. SAR Robotics – Areas of Application

Of course, one can imagine a wide area of possible robotics applications in the Search and Rescue domain. It is, however, obvious that many of these ideas require technology far beyond the current state of the art. The choice presented in this work rather addresses applications already in use or expected for the near future. It roughly adapts the use cases discussed in the upcoming EU Inducement Price for Robotics for Humanitarian Assistance and Search & Rescue.

### A. Environment Monitoring

Almost all aspects of the environment can be monitored by using robotics technology as the main means of mapping an area. This includes non-SAR applications as well, ranging from crop monitoring to pollution control to water quality monitoring. But also in post-disaster scenarios various parameters of the environment have to be steadily checked, e.g. chemical, biological or radiological hazards after accidents in industrial facilities or, worse, after terrorist attacks. Measuring such possible threats often cannot be conducted by human personnel simply because it is too dangerous.

### B. Inspection

Another important application is the long-term stability assessment of partially wrecked structures. Using high-precision 3D laser scanners unmanned vehicles can monitor if it is still safe for humans to enter and work in the wrecked areas. As a pre-disaster application, autonomous inspection systems have the potential to reduce costs and increase the thoroughness of inspections of buildings, such as chimneys, historic buildings, bridges and tunnels. The use of autonomous systems may provide better early warning of issues within structures. Robots have also been deployed in the nuclear industry for internal reactor inspection reducing human risk levels.

### C. Search and Rescue

Robotic systems have been deployed to carry out search operations after natural disasters. The uptake of search systems has, to date, been low and the potential to deploy robot systems requires further investigation and exploitation. Examples are the survey of over 10,000 m$^2$ of submerged areas in less than 16 hours in the water, finding over 100 major objects to be removed during the 2011 Japan Earthquake and Tsunami, and the use of robots to explore the basement during the search operations at the World Trade Centre (9/11) [4]. The ideal scenario would be the scaling up to high level collaborative autonomous systems capable of scanning large areas during SAR operations. These systems should create maps of spaces and identify voids in collapsed buildings improving search and rescue outcomes. The use of multiple robots providing coordinated search in unknown and dynamic environments that are typical of disaster zones could provide enhanced safety to rescue workers and increase the likelihood of discovering victims and identifying threats and hazards. The collaboration between human workers and robots is promising in this scenario but requires communication and control in emergency situations such as 9/11 [5]. There are potentially significant safety gains with the use of a tele-operated semi-autonomous robot during search operations, used to enter buildings and carry out search and possibly rescue tasks. The robot should be able to reach spaces and regions of a building that a human operator may not, and it may be able to move faster and with significantly lower risk. On finding a person its internal map of the space can be used to plot the optimal route to affect a recovery. In more advanced systems the robot may be able to provide basic medical assessments and basic medication increasing survivability, even in a simple way such as delivering water or pain relief [6].

## III. SAR Robotics – Demands and Requirements

Looking at the literature one can find a lot of work regarding SAR-specific demands and requirements for robot systems, which are important and necessary for a successful deployment; see for example [3, 7, 8]. However, most of these articles address special SAR sub-domains, like Urban Search & Rescue (USAR) [7] or chemical, biological, radiological, nuclear and explosive (CBRNE) incident response [8]. The following list is a compilation of requirements common to the SAR field as a whole.

### A. Search-specific Requirements

*1) Navigation and mapping*

Autonomously produce a 3D map of the interior and exterior of the scene, which can be communicated to operators and enable structural assessments that inform of the safety of a building.

*2) Casualty identification*

A robot must be able to autonomously identify trapped hidden casualties in the environment. Possible technologies include visual, audio or heat sensing methods.

### B. Rescue-specific Requirements

*1) Communication*

Communication is of fundamental importance to robot search and rescue. Regardless of the degree of autonomy, a continuous communication link is required between the search and rescue robot (or robots) and their operators. This operator-robot communication link needs to be

- duplex, in order to allow the operator to send command/control data while receiving video, sensor or status data from the robot;
- continuous, with low latency, in order to allow smooth uninterrupted control or supervision of the robot;
- secure and reliable, to avoid unintended interference or signal loss from other radio sources or as a result of environmental factors;
- normally high-bandwidth, to allow for streaming real-time video.

Under all circumstances the system should be able to maintain communication with the operator or at least with other members of the team, i.e. robots or human personnel. The robot should be able to provide audio-visual communication between operator and trapped persons that allows for basic diagnosis by an emergency responder.

*2) Support*

The ability of the solution to provide assistance to the casualty is required. At least one robot in the team must be able to deliver water and a payload representing basic medical supplies to a casualty.

*3) Remote mobile manipulation*

The unmanned system should be able to manipulate and remove debris, and to open a closed door.

*C. General Requirements*

*1) Time of operation/battery life*

The system's working time must be sufficient for typical application scenarios. In [3], for example, for UGVs a working time of more than one hour is suggested, which also fits well to the typical scenario time of real-life robotic competitions (like ELROB or the DARPA Robotics Challenge). For UAVs a shorter operational time is acceptable. The robot system must be set up in as short a time as possible.

*2) User interface/operation*

The system should have a user interface which considers the needs of the intended users, typically SAR first response teams. In particular easiness of use and reduced training is important for operators.

*3) Safety*

No part of the robotic system should represent a risk to operators, rescue personnel or casualties.

*4) Portability*

It must be possible to easily transport the whole system to a disaster site.

IV. CURRENT STATE OF THE ART

This section thoroughly pictures the current situation of robotic systems and sub-systems in use in today's SAR and disaster response scenarios.

*A. Platforms*

The number of companies that manufacture commercially available UGVs for the SAR domain in mentionable quantities is very limited. Most of the professional robotic systems that are deployed in the field come from the bomb disposal domain.

Due to this fact, most of these robots have a very similar appearance. Usually the platforms are propelled by tracks and run on batteries. Most of them are lead acid battery-based while some of the newer models use some form of lithium-based batteries. The vast majority of the systems have manipulators for handling the explosive device. The systems are usually not very fast and the stair-climbing capabilities require quite a bit of training for the operator.

Additionally, a lot of concept studies and experimental systems can be found, but all far from being ready for deployment. Regarding aerial systems, often commercial quad- or hexa-copters are used to generate a visual scene overview or, sometimes, to generate laser-based 3D maps of the surroundings, like e.g. in [9]. Generally, all these systems suffer from a very limited maximum time of operation.

*B. Obstacle Avoidance and Path Planning*

A search and rescue robot typically requires short- or medium-range proximity sensors for obstacle avoidance, such as infrared return-signal-intensity or ultrasonic- or laser-based time-of-flight systems. The most versatile and widely used device is the 2D or 3D laser range finder, which can provide the robot with a set of radial distance measurements and hence allows the robot to plan a safe path through obstacles. For a comprehensive review of motion planning and obstacle avoidance in mobile robots see [10].

*C. Localisation*

All but the simplest search and rescue robots will require sensors for localisation that enable the robot to estimate its own position in the environment. If external reference signals are available – such as fixed beacons so that a robot can use radio trilateration to fix its position relative to those beacons, or a satellite navigation system such as GPS – then localisation is relatively straightforward. Otherwise, a robot will typically make use of several sensors including odometry, an inertial measurement unit (IMU) and a magnetic compass, often combining the data from all of these sensors, including laser-scanning data, to form an estimate of its position. Simultaneous localisation and mapping (SLAM) is a well-known approach which typically employs Kalman filters to allow one or more robots to both fix their position relative to observed landmarks and map those landmarks with increasing confidence as the robots move through the environment [11, 12].

*D. Object Detection*

Vision is often the sensor of choice for object detection in laboratory experiments for SAR robots. If, for instance, the object of interest has a distinct colour that stands out in the environment then standard image processing techniques can be used to detect it and steer towards the object. However, if the environment is visually cluttered, unknown or poorly illuminated then vision becomes problematical. Alternative approaches to object detection include, for example, artificial odour sensors: Hayes et al. demonstrated a multi-robot approach to localisation of an odour source [13]. An artificial whisker modelled on the Rat mystical vibrissae has been demonstrated [14]; such a sensor could be of particular value in dusty or smoky environments.

*E. Locomotion*

The means of physical locomotion for a search and rescue robot can take many forms and clearly depend on the environment in which the robot is intended to operate. Ground robots typically use wheels, tracks or legs, although wheels are predominantly employed in proof-of-concept or demonstrator SAR robots. Whatever the means of locomotion, important principles that apply to all search and rescue robots are that a robot must be able to move with sufficient stability for the object detection sensors to be able to operate effectively; and it must be able to position itself with sufficient precision and stability. These factors place high demands on a search and rescue robot's physical locomotion system, especially if the robot is required to operate in soft or unstable terrain.

*F. Object Manipulation*

The manipulation required by a search and rescue robot is clearly dependent on the form of the search object of interest and the way the object presents itself to the robot as it approaches. The majority of search and rescue experiments or demonstrations have simplified the problem of object manipulation by using objects that are, for instance, always the right way up so that a simple gripper mounted on the front of the robot is able to grasp the objects with reasonable reliability. However, in general a search and rescue robot would require the versatility of a robot arm (multi-axis manipulator) and general-purpose gripper (hand) such that – with appropriate vision sensing – the robot can pick up the object regardless of its shape and orientation. These technologies are well developed for tele-operated robots used for remote inspection and handling of dangerous materials or devices [15].

*G. Communication*

The demanding communication requirements listed in the former section are often tested to the limit in real-world emergency scenarios. Both wired and wireless communication links are employed in SAR robots. Wireless communication is the preferred mode, although reliable wireless communication can be very problematical when search and rescue robots must be deployed in buildings, metal structures or under high levels of radiation. Wired (cable) connections suffer a different set of problems, because of the problems of managing cable spooling and run-out and the need to avoid cable snagging in the environment, or on the robot itself.

For an introduction to communications and networking for teleoperation see [16]. Wireless local area network (WLAN) technology is highly appropriate for terrestrial robot systems, and the advantages of the technology (including bandwidth and reliability) are sufficient to justify the proposed use of intermediate robots acting as wireless relays between the operator and the SAR robot (see for instance [17]). Future multi-robot search and rescue systems can take advantage of the fact that a spatially distributed team of wireless networked robots naturally forms an ad hoc network, which – providing the team maintains sufficient connectivity – allows any robot to communicate with any other via multiple hops. As long as the operator maintains connection with one of the robots, a multi-hop multi-path network connection is then maintained with all robots [18].

*H. Human-Robot Interfaces*

Search and rescue robots by definition need to work as part of human rescue teams and therefore – whatever the level of autonomy – need a human-robot interface (HRI). The design of the HRI is of great importance. A well-designed human-robot interface significantly increases a search and rescue robot's usability and this, in turn, is likely to lead to greater deployment and value to the rescue team.

The essential ingredients to be found in current SAR human-robot interface are:

- the means to control the robot's locomotion, i.e. joystick or equivalent;
- the means to control the robot's actuator, i.e. robot arm, gripper or equivalent device;
- video displays to see what the robot's camera(s) are seeing – and to control camera functions such as pan, tilt and zoom;
- video displays or readout devices, to allow monitoring of key environmental measurements, such as temperature, pressure, humidity, radiation level or hazardous gases;
- video displays or readout devices, to allow monitoring of a robot's status, including battery level, the robot's attitude, altitude/depth, location and nearby objects, etc.

As Murphy et al. conclude in [19], the human-robot interface is a major challenge in rescue robotics that 'has been declared to be an exemplar domain within human-robot interface'.

*I. Autonomy and Tele-Operation*

A robot's autonomy describes the degree to which it can make decisions about its next possible action without human intervention. Autonomy thus falls on a spectrum, from fully tele-operated robots – robots with zero autonomy – at one end, and fully autonomous robots – robots capable of completing their mission from start to end without human intervention – at the other. Search and rescue robots might, in principle, be found anywhere on this spectrum of autonomy, but in practice they are either tele-operated or semi-autonomous. Additionally, all robots need some degree of human supervision; we refer to this as supervised autonomy.

A fully tele-operated robot is one in which a human operator controls every function of the robot directly via some data link [20]. The data link provides a continuous connection between the robot and its operator's control station. Full tele-operation places a considerable burden on the human operator, since he needs to continuously watch and interpret the video feed and provide continuous control of motors while steering around obstacles, navigating the terrain, etc. Semi-autonomous operation, also referred to as supervisory control, reduces this burden. For an overview of telerobotics see [21].

A semi-autonomous robot is one in which some, often low-level, functions can be left to the robot while high-level control remains with the human operator. A common approach to semi-autonomous operation – especially in UAVs – is for the human pilot to set a target destination (waypoint) then leave the low-level control required to reach the destination to the aircraft's autopilot. The same semi-autonomous approach is perfectly possible for ground SAR robots, although the autonomous control functions may need to be more complex to enable the robot to, for instance, safely navigate rough terrain or steer around obstacles.

Another, higher, level of semi-autonomous operation allows a robot to search some bounded area for objects of interest – then perhaps halt and alert its human operator when an object is found. This mode would be most appropriate if the robot is searching for survivors or, say, some single critical object. Another mode might require a robot to autonomously search the entire area, find and localise each object of interest, then – once the area has been covered – halt and provide its operator with a map marking the position of the found objects. Such levels of autonomy, however, so far only exist in experimental systems.

## V. FUTURE DIRECTIONS FOR SAR ROBOTICS

As Murphy et al. make clear, search and rescue robotics is an emerging field, which has a long way to go before it reaches its full potential [19]. Almost every advance in intelligent autonomous robotics has the potential to benefit search and rescue robotics. In this section we outline a number of directions that, either individually or jointly, could lead to significantly more capable SAR robots in the medium and long-term future.

*A. Heterogeneous Multi-Robot Multi-Domain SAR Robots*

Search and rescue is clearly a task that lends itself to multi-robot systems and, even if a single robot can accomplish the task, SAR should – with careful design of strategies for cooperation – benefit from multiple robots. The most significant advantage of multiple robots is the ability to cover a much larger area and hence reduce the time to find survivors or critical hazards. Another benefit is gained by combining the advantages of robots in different domains; for example a flying robot providing a birds-eye view of the scene to guide a land robot's search.

At the time of writing there are no known examples of multi-robot systems deployed alongside real-world SAR teams. The principle reason for this is the difficult problem of controlling and coordinating a multi-robot team. Tele-operating a single robot can be challenging – so tele-operating a whole team is probably beyond even the most skilled human operators.

The solution to this problem will be found in a combination of greater individual robot autonomy and advanced human-multi-robot interfaces. Consider autonomy; there are two paradigms for the control and coordination of multiple robots: multi-robot systems (MRS) or swarm robotics systems. Multi-robot systems are characterised as centrally controlled, whereas in swarm systems control is distributed and decentralised. It is, however, an open question when such swarm robotics systems will reach a deployable state.

*B. Dynamic Autonomy in SAR*

Consider the situation in which a semi-autonomous SAR robot is searching inside a structure. If the structure contains – as is likely – unknown hazards, then it is possible that the robot encounters a problem that is too difficult for its intelligent search capability to cope with. Ideally we would like the robot to be able to detect when its semi-autonomous capability has been exceeded, halt (safely), then 'ask' its human operators to resume control. We describe this as dynamic autonomy. Baker and Yanco outline the potential for dynamic autonomy in an urban rescue scenario [22]; Schermerhorn and Scheutz investigate dynamic autonomy in human-robot teams [23]. Dynamic autonomy would be a significant advance for SAR robots, but is not straightforward to implement, both because of the complex human factors and because it requires that the robot is able to assess the level of danger posed by hazards before it becomes irrecoverably stuck or damaged.

*C. Immersive Telepresence*

After more than 20 years in development it now appears that Virtual Reality (VR) headsets are set to become a practical, workable proposition; the low-cost Oculus Rift VR headset, for example, integrates 3D gyros, accelerometers and a magnetometer – and claims to reduce latency to very low levels. Of course the primary market for VR headsets is likely to be entertainment, including video games. VR could, however, revolutionise the human-robot interface for tele-operated robots as well.

Consider a tele-operated robot with a pan-tilt camera linked to the remote operator's VR headset, so that every time she moves her head to look in a new direction the robot's camera moves in sync; so the operator sees (and hears) what the robot sees and hears in immersive high definition stereo. Of course the reality experienced by the robot's operator is real, not virtual, but the head mounted VR technology is the key to making this work. Reis and Ventura describe work at the Intelligent Robot and Systems Group, IST Lisbon, in which a stereo camera with pan-tilt mechanism mounted on a tracked mobile robot is coupled to a head-mounted display with head tracker system [24].

With the addition of haptic gloves for control, the robot's operator would have a highly intuitive and immersive interface with the robot. The illusion of 'being in' the robot could well provide the operator with much more natural sense of the robot's position and its immediate surroundings. The haptic gloves would provide the operator with the ability to, for instance, move the robot's arm and gripper simply by moving her own arm and hand. Such a system is, for example, presented in [25], showing that it easily outperforms classical manipulator steering devices.

*D. Novel Hardware Designs for SAR Robots*

The design of current SAR robots, and in particular their morphology and locomotion, has its origins in vehicle design. Ground SAR robots are generally wheeled or tracked vehicles following a conventional pattern; SAR UAVs are aircraft without pilots. However, the emergence, in the last decade, of bio-inspired and bio-mimetic robotics is leading to new animal-like hardware developments. For reviews of bio-inspired robots – including humanoid robots – see for example [26]. Although none have yet been deployed into SAR teams or emergency services it seems likely that they will be soon.

*1) Snake robots*

Using neither legs nor wheels, snake-like robots have been proposed for navigating terrain, small enclosed spaces or pipes, which would be impossible for conventional robots. Probably the most developed prototypes are the Japanese Soryu and ACS snake-like robots [27]. Another example is the snake-like hyper-redundant robot (HRR) for urban search and rescue from the Biorobotics and Biomechanics Lab of the Technion Israel Institute of Technology [28]; this robot has 14 serially chained actuated segments, each of which is capable of supporting the entire robot structure.

*2) Legged robots*

Although not especially designed for SAR the Boston Dynamics BigDog robot is perhaps the best-known example of an advanced quadrupedal robot designed for rough terrain. However, future SAR robots might need to be legged, in order to achieve the same versatile mobility as humans, horses or dogs. With an explicit target of SAR applications are the legged quadrupedal robots HyQ and StarETH. HyQ is a hydraulically actuated quadruped developed at the IIT's Department of Advanced Robotics, and StarETH is a quadruped based on series elastic actuation developed at ETH Zurich's Agile and Dexterous Robotics Lab. Additionally, these two labs have launched a collaborative project named AGILITY to further improve locomotion.

### 3) Humanoid rescue robots

Although not specifically designed for SAR tasks, the ATLAS humanoid robot has been provided to participants of the SAR-related DARPA Robotics Challenge (DRC). Designed by Boston Dynamics ATLAS is a hydraulically actuated humanoid robot of height 1.88m and weight 155kg. The robot has 28 actuated degrees of freedom, and requires a tethered connection for power, cooling (water) and wired communications.

It is supposed that humanoid robots would, in SAR situations, have the advantage of being able to use tools and devices designed for humans, including vehicles, and move more readily through human environments. However, whether a humanoid robot (even something smaller, lighter and more autonomous than ATLAS) would actually outperform a well-designed conventional SAR robot remains an open yet interesting question.

## VI. CONCLUSION

This work described typical areas of Search and Rescue (SAR) applications: environment monitoring, disaster inspection and the actual SAR task. Additionally, it thoroughly pictured SAR-specific demands and requirements on the various components of a robotic system, addressing e.g. navigation and mapping, communication, battery life and time of operation, manipulation capabilities, or user interface design. A state-of-the-art overview of robotic systems and sub-systems currently in use in SAR and disaster response scenarios has been given, as well as an outlook on expected future developments in this domain. In summary, all this should illustrate that the idea of taking SAR applications as applied benchmark for Field Robotics (FR) systems will surely remain valid for the next years if not decades.

REFERENCES

[1] R. Snyder, Robots assist in search and rescue efforts at WTC, IEEE Robotics and Automation Magazine, vol. 8, 2001, pp. 26-28.

[2] G. De Cubber, D. Doroftei, D. Serrano, K. Chintamani, R. Sabino and S. Ourevitch, The EU-ICARUS project: developing assistive robotic tools for search and rescue operations, Proceedings of the IEEE International Workshop on Safety, Security & Rescue Robotics (SSRR), Linköping, 2013, pp. 1-4.

[3] R. R. Murphy, Disaster Robotics, 1st ed., MIT Press, Cambridge, 2014.

[4] CRASAR, Documented Use of Robots for Disasters (2015) [online], Available from: http://crasar.org/disasters/, accessed 22nd Sept 2015.

[5] J. Casper and R. R. Murphy, Human-robot interactions during the robot-assisted urban search and rescue response at the World Trade Center, , IEEE Transactions on Systems, Man, and Cybernetics, Part B: Cybernetics, vol. 33(3), 2003, pp. 367-385.

[6] R. R. Murphy, D. Riddle and E. Rasmussen, Robot-Assisted Medical Reachback: A Survey of How Medical Personnel Expect to Interact with Rescue Robots, Proceedings of the 2004 IEEE International Workshop on Robot and Human Interactive Communication, Okayama, 2004, pp. 301-306.

[7] J. Casper, M. Micire and R. R. Murphy, Issues in Intelligent Robots for Search and Rescue, in: Unmanned Ground Vehicle Technology II - Proceedings of SPIE, vol. 4024, 2000, pp. 292-302.

[8] C. M. Humphrey and J. A. Adams, Robotic tasks for chemical, biological, radiological, nuclear and explosive incident response, Advanced Robotics, vol. 23(9), 2009, pp. 1217-1232.

[9] G.-J. M. Kruijff, V. Tretyakov, T. Linder, F. Pirri, M. Gianni, P. Papadakis, M. Pizzoli, A. Sinha, E. Pianese, S. Corrao, F. Priori, S. Febrini and S. Angeletti, Rescue robots at earthquake-hit Mirandola, Italy: A field report, Proceedings of the IEEE International Symposium on Safety, Security, and Rescue Robotics (SSRR), College Station, 2012, pp. 1-8.

[10] J. Minguez, F. Lamiraux and J.-J. Laumond, Motion Planning and Obstacle Avoidance, Springer Handbook of Robotics, B. Siciliano and O. Khatib (eds.), Spinger, 2008, pp. 827-852.

[11] M. W. M. G. Dissanayake, P. M. Newman, H. F. Durrant-Whyte, S. Clark and M. Csorba, A solution to the simultaneous localization and map building (SLAM) problem, IEEE Transactions on Robotics and Automation, vol. 17(3), 2001, pp. 229-241.

[12] S. Thrun and J. J. Leonard, Simultaneous Localization and Mapping, Springer Handbook of Robotics, B. Siciliano and O. Khatib (eds.), Spinger, 2008, pp. 871-890.

[13] T. Hayes, A. Martinoli and R. M. F. Goodman. Distributed odor source localization. IEEE Sensors, Special Issue on Artificial Olfaction, vol. 2(3), 2002, pp. 260-271.

[14] M. J. Pearson, A. G. Pipe, C. R. Melhuish, B. Mitchinson and T. J. Prescott. Whiskerbot: A robotic active touch system modeled on the rat whisker sensory system, Adaptive Behavior, vol. 15, 2007, pp. 223-240.

[15] T. Schilling (ed.), Telerobotic Applications, John Wiley and Sons, 1999.

[16] D. Song, K. Goldberg and N. K. Chong, Networked Telerobots, Springer Handbook of Robotics, B. Siciliano and O. Khatib (eds.), Spinger, 2008, pp. 758-771.

[17] Ö. Çayırpunar, B. Tavlı and V. Gazi, Dynamic Robot Networks for Search and Rescue Operations, Proceedings of the RISE 2008 Workshop on Robotics for Risky Environments, Benicàssim, 2008.

[18] S. Hauert, S. Leven, M. Varga, F. Ruini, A. Cangelosi, J.-C. Zufferey and D. Floreano, Reynolds flocking in reality with fixed-wing robots: Communication range vs. maximum turning rate. Proceedings of the IEEE/RSJ International Conference on Intelligent Robots and Systems, San Francisco, 2011, pp. 5015-5020.

[19] R. R. Murphy, S. Tadokoro, D. Nardi, A. Jacoff, P. Fiorini, H. Choset and A. M. Erkmen, Search and Rescue Robotics, Springer Handbook of Robotics, B. Siciliano and O. Khatib (eds.), Spinger, 2008, pp. 1151-1173.

[20] A. F. T. Winfield, Future directions in tele-operated robotics, Telerobotic Applications, T. Schilling (ed.), John Wiley and Sons, 1999, pp. 147-162.

[21] G. Niemeyer, C. Preusche and G. Hirzinger, Telerobotics, Springer Handbook of Robotics, B. Siciliano and O. Khatib (eds.), Spinger, 2008, pp. 741-757.

[22] M. Baker and H. A. Yanco, Autonomy mode suggestions for improving human-robot interaction, Proceedings of the IEEE International Conference on Systems, Man, and Cybernetics, The Hague, 2004, pp. 2948-2953.

[23] P. Schermerhorn and M. Scheutz, Dynamic robot autonomy: investigating the effects of robot decision-making in a human-robot team task, Proceedings of the International Conference on Multimodal Interfaces, Cambridge, 2009, pp. 63-70.

[24] J. Reis and R. Ventura, Immersive robot teleoperation using a hybrid virtual and real stereo camera attitude control, Proceedings of the 18th Portuguese Conference on Pattern Recognition, Coimbra, 2012.

[25] B. Brüggemann, D. Wildermuth and F. E. Schneider, Search and Retrieval of Human Casualties in Outdoor Environments with Unmanned Ground Systems – System Overview and Lessons Learned from ELROB 2014, Proceedings of the 10th Conference on Field and Service Robotics (FSR), Toronto, 2015.

[26] J.-A. Meyer and A. Guillot, Biologically Inspired Robots, Springer Handbook of Robotics, B. Siciliano and O. Khatib (eds.), Spinger, 2008, pp. 1395-1422.

[27] K. Hatazaki, M. Konyo, K. Isaki, S. Tadokoro and F. Takemura, Active Scope Camera for Urban Search and Rescue, Proceedings of the 2007 IEEE/RSJ International Conference on Intelligent Robots and Systems, San Diego, 2007, pp. 2596-2602.

[28] A. Wolf, H. Choset, H. B. Brown and R. Casciola, Design and Control of a Mobile Hyper-Redundant Urban Search and Rescue Robot, Int. Journal of Advanced Robotics, vol. 19(8), 2005, pp. 221-248.